\pdfoutput=1
\documentclass[11pt]{article}
\usepackage[preprint]{acl}%

\usepackage{times}
\usepackage{latexsym}
\usepackage[T1]{fontenc}
\usepackage[utf8]{inputenc}
\usepackage{microtype}
\IfFileExists{inconsolata.sty}{\usepackage{inconsolata}}{}%
\usepackage{graphicx}
\usepackage{amsmath,amssymb}
\usepackage{booktabs}
\usepackage{array}
\usepackage{xcolor}
\usepackage{enumitem}

\newcommand{\gn}{G0.5}
\newcommand{\vv}[1]{v_{\text{#1}}}
\newcommand{\tok}[1]{\texttt{\detokenize{#1}}}

\title{Says Left, Reaches Right: \\ Language and Action Diverge in a Hybrid VLA}

\author{Aadi Palnitkar \\ University of Maryland \And Dmytro Kurdydyk \\ Independent Researcher}

\makeatletter
\newcommand{\panellabel}[2]{{\edef\@currentlabel{\thefigure#1}\label{#2}}}
\makeatother
\begin{document}
\maketitle

\begin{abstract}
Interpretability methods can now read what a model will say before it says it. The Jacobian lens (J-lens) reads
representations that its authors argue constitute a global workspace in LLMs. However, whether this workspace extends
to physical action remains unclear. If a robot policy shares a workspace for words and actions, reading and editing its
verbal representations should let us predict and steer its behavior. We test this in Galaxea \gn{}, a
vision-language-action model (VLA) that writes a plan, an arm token and discrete action
tokens with shared weights in one decoding stream. Despite this architectural unification, editing what the lens reads
changes the plan's stated side in 88\% of plans naming one, yet redirects none of 35 reaches. By contrast, the recurrent
memory of a prompt naming the other hand redirects 60\% once the plan and arm tokens are forced to that side.
Activation patching reveals that the plan's words depend primarily on attention, while action tokens depend on
recurrent GatedDeltaNet (GDN) blocks. Notably, this verbal--motor dissociation is absent in \gn{}'s base model on a text
task. Together, these findings challenge the assumption that reading what a robot policy would say reveals what it will
do.
\end{abstract}

\begin{figure*}[t]
  \centering
  \includegraphics[width=\textwidth]{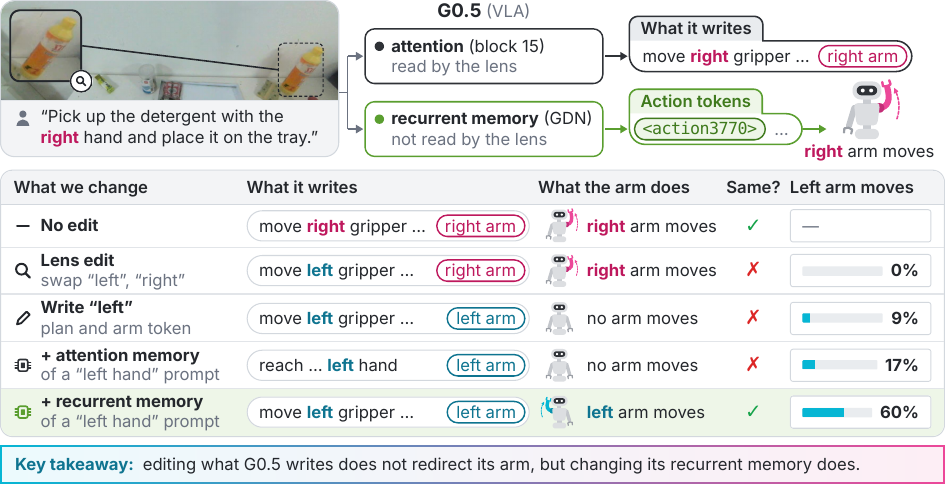}
  \caption{\textbf{Plan text vs.\ decoded arm movement.} The top row shows the pipeline on a public demonstration
  (Appendix~\ref{app:demo}). Each row below edits one study sample (right hand, paper-roll cover) and shows what \gn{} writes
  and which arm moves. The bar gives the share of the 35 reaching samples in which the left arm moves.}
  \label{fig:teaser}
\end{figure*}

\section{Introduction}

\looseness=-1 Whether what a model says reflects the computation behind its behavior is a central question for NLP, because chains of thought can be unfaithful to the reasons for an answer~\citep{turpin2023language,lanham2023measuring}. The Jacobian
lens (J-lens) of \citet{gurnee2026workspace} instead looks inside the model and maps a block's activations to the
tokens they are poised to make the model say. Its authors argue that these verbalizable representations form a
global workspace, a term from cognitive neuroscience~\citep{baars1988cognitive,dehaene1998neuronal} for a
limited-capacity stage whose contents are broadcast to many specialized processes at once, which is thought to make
information consciously accessible, reportable and usable for control. In LLMs, \citet{gurnee2026workspace} argue that it works as a working memory, whose contents can be deliberately held and carry silent reasoning, and
that the J-lens lets us view it. Reading this workspace directly would not depend on the chain of thought, which matters for safety. Yet the
workspace is selective, since swapping two
concepts' lens coordinates flips what a model \emph{reports}, such as a passage's language, but not automatic
processing, such as continuing it.

\looseness=-1 Embodied agents make the question concrete, because their motion can be measured apart from their words and their physical actions raise new safety concerns. Reasoning vision-language-action (VLA)
policies write a plan before they act~\citep{zawalski2024robotic}. Editing the plan can redirect the
actions~\citep{trinh2026altered}, yet actions can also miss the stated reasoning~\citep{foutter2026do}.
If words and actions share a workspace anywhere, it should be in \gn{}~\citep{galaxea2026g05}. Unlike VLAs that keep goals and motor programs
in separate pathways~\citep{grant2026not}, \gn{} treats language and actions as
tokens of one vocabulary from one backbone, sharing the decoder, context and objective, so a single lens assigns a
direction to every output, and its authors hypothesize that the action tokens attend to the reasoning.

\looseness=-1 We therefore apply the J-lens to \gn{} and ask whether language and action share a workspace: does editing
what the lens reads redirect both the plan and the arm, as changing the instruction does?

\looseness=-1 In \gn{}, language and action do not share one, even though it has no separate pathway for actions by design. Its words (the plan and, partly,
the token that selects the arm) draw on attention, and its action tokens on recurrent GatedDeltaNet (GDN) blocks
(Figure~\ref{fig:teaser}, \S\ref{sec:routes}). Editing what the lens reads rewrites the plan's side in 88\% of plans
naming one but redirects none of 35 reaches, a verbal--motor dissociation, while the recurrent memory of a prompt
naming the other hand redirects 60\% once the plan and arm tokens are forced to that side (\S\ref{sec:behaviour}).
Hybrid language models, including Qwen3.5, likewise report an instruction through attention and follow it through
their recurrent state~\citep{afendulev2026attention}. Yet on a text task, Qwen3.5-2B-Base, probably \gn{}'s
initialization, follows a forced written plan in 96\% of prompts, while \gn{}'s reach follows it in 6\%
(Figure~\ref{fig:contrast}). We contribute:
\begin{itemize}[leftmargin=*,itemsep=1pt,topsep=2pt,parsep=0pt]
  \item \textbf{The first J-lens analysis of a VLA}, to our knowledge, finding that the lens first reads a concept at distant positions at one attention block (\S\ref{sec:readable}).
  \item \textbf{A verbal--motor dissociation}, in which words draw mainly on attention and action tokens on recurrence,
  and lens edits change the words without redirecting the arm, which the recurrent memory moves (\S\ref{sec:results}).
  \item \textbf{A base-model contrast}, in which Qwen3.5-2B-Base, \gn{}'s architecture without robot training, follows its
  written plan on a text task (\S\ref{sec:behaviour}).
\end{itemize}

\section{Background}
\label{sec:background}

\looseness=-1 The J-lens~\citep{gurnee2026workspace} refines the logit~\citep{nostalgebraist2020logit} and
tuned~\citep{belrose2023tuned} lenses with an averaged Jacobian, with follow-ups across models and
readouts~\citep{wang2026looped,wang2026residual,yan2026short,gong2026first,he2026sparse}.
We mainly use the J-lens, but the logit lens gives the same jump at block 15 and similar swap counts
(\S\ref{sec:readable}, Appendix~\ref{app:more}), so our findings concern the directions in the residual stream that lenses read, not only the J-lens's transport.

\looseness=-1 In a
hybrid LM, attention makes a latent variable readable to lenses~\citep{mazaheri2026gathered}, and in
recurrent hybrids, recurrent layers prepare the keys that attention matches in induction~\citep{cheng2026token}.

\looseness=-1 In VLAs, actions can miss a correct plan or instruction~\citep{wu2025do,jin2026stage,foutter2026do}, reasoning can
be epiphenomenal or causal~\citep{nguyen2026vladrivebench,yin2025deepthinkvla}, and corrections to it seldom move
actions~\citep{karnik2026reasoning}.
VLA internals have also been probed and
steered \citep{haon2025mechanistic,swann2026sparse,buurmeijer2026observing,grant2026not,chan2026steering}.
Instructions stay decodable and shift an action lens~\citep{jiang2026encoded}, yet often fail to
control action, and in VLAs with separate pathways, mismatched goals stall~\citep{grant2026not}. However, none of
these VLA studies uses the J-lens or splits a concept's routes by block type.

\begin{figure*}[t]
  \centering
  \includegraphics[width=\textwidth]{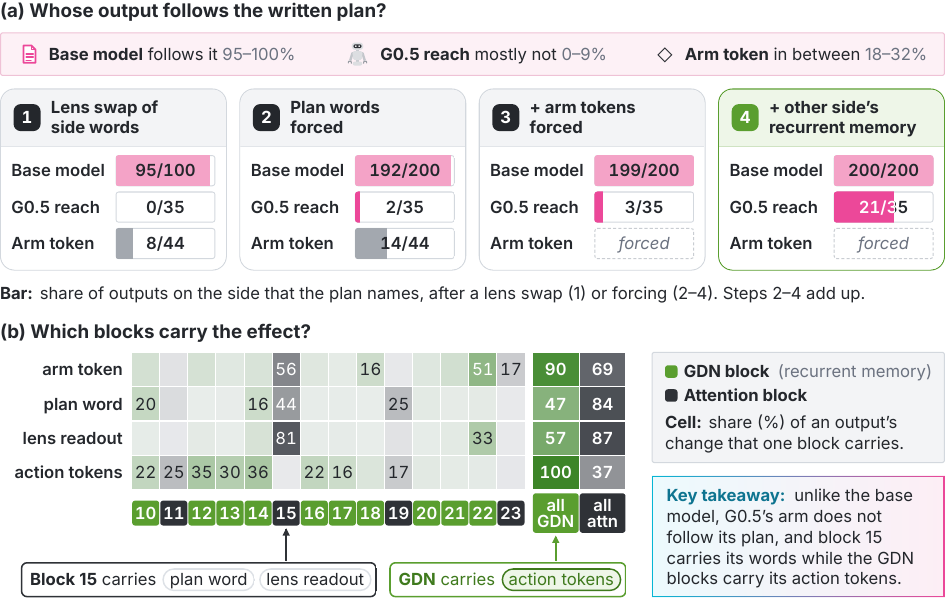}
  \caption{\textbf{The written plan decides the base model's output but not \gn{}'s reach (a). Attention block 15 carries
  the words, and the GDN blocks the action tokens (b).} (a)~Share of outputs on the plan's side: Qwen3.5-2B-Base on a
  text task (its sign in the step-1 pilot, then digit codes in \gn{}'s format; Appendix~\ref{app:base}), \gn{}'s reach (of 35) and
  first arm token (of 44); step 4 uses the 18 GDN blocks' memory. (b)~Share of the counterfactual effect reproduced by patching one block's, or all GDN or attention blocks', token-mixing output
  ($\ge$15\% shown; blocks 0--9: Appendix~\ref{app:more}).}
  \label{fig:pair}
  \panellabel{a}{fig:contrast}\panellabel{b}{fig:patching}
\end{figure*}

\section{Setup}
\label{sec:setup}

\paragraph{Model and output format.}\looseness=-1 The backbone of \gn{}, initialized from Qwen3.5-2B, has 24 blocks (0--23). Their
token-mixing layer is full attention in blocks 3, 7, 11, 15, 19 and 23 and GatedDeltaNet (GDN), linear attention
with a recurrent state~\citep{yang2025gated}, in the other 18. Given three camera images, the robot state and an
instruction, it greedily writes a plan (``\tok{Subtask:} reach for the door handle with the right gripper''), then per moving arm an \emph{arm token} such as \tok{<right_control_0>}, eight
action tokens, a second arm token and eight action tokens.
A codec decodes them into an action chunk of joint positions (not a rollout).

\paragraph{Samples and arm outcomes.}\looseness=-1 We use saved generations on 120 tasks of the Galaxea Open-World
Dataset~\citep{galaxea2025g0}, one observation each. In 75 \emph{eligible} samples the instruction names one
hand, the arm tokens one arm. An arm \emph{reaches} if its largest joint range exceeds 0.3\,rad. One arm reaches in
35 of the 75 clean runs, always the instructed one. Such a sample is \emph{redirected} if only the other arm reaches under
intervention and \emph{stopped} if no arm does. As every clean reach uses the instructed arm, a redirect is
unambiguous, and the low reach rate limits the sample size and the behavior covered. Redirect and stop rates are over these 35, and rates of flipped plans over the 51 plans that name a side (counts in Appendix~\ref{app:stats}).

\paragraph{J-lens definition and fitting.}\looseness=-1 The lens of block $\ell$ averages over calibration prompts and source positions $t$ the summed
Jacobian $J_\ell = \mathbb{E}_t\,[\sum_{t' \ge t} \partial h_{23,t'} / \partial h_{\ell,t}]$. With the final RMSNorm gain $g$ folded into
unembedding row $u_w$, token $w$'s lens vector is $v_w = J_\ell^\top (g \odot u_w)$ and its lens logit at $h$ is $v_w
\cdot h / \mathrm{rms}(J_\ell h)$. Lenses fitted on action tokens (action lenses) collapse (Appendix~\ref{app:lens}), so we
fit text lenses with the reference code on WikiText, a main lens on 10 prompts (enough to beat the logit lens
in~\citealp[App.~A.7]{gurnee2026workspace}) and a replication lens on 128. We rely on lens
directions, chiefly the side contrast $\vv{left}-\vv{right}$, whose cosine across disjoint prompt sets is
0.93--1.00 in blocks 9--21 (Appendix~\ref{app:lens}).

\paragraph{Intervention types.}\looseness=-1 The \emph{counterfactual instruction} replaces the side word by the other one
(\mbox{`` right''$\to$`` left''}), or an object noun by another. In \emph{teacher-forced} replays every generated
token is kept clean, and $d = h_{\text{cf}} - h_{\text{clean}}$ is the counterfactual change. In \emph{generation} the
policy decodes natively, optionally with forced tokens or part of its prompt memory (GDN states or attention keys and values) from
another prompt (a memory swap, as in~\citealp{afendulev2026attention}).
The \emph{J-lens swap}~\citep{gurnee2026workspace} clamps the two words' lens coordinates $c=V^{+}h$
($V=[v_s\,v_t]$) to $c_0+\alpha(\sigma(c_0)-c_0)$, where $c_0$ is the
unedited run's and $\sigma$ swaps the two, at blocks 9--21 (their workspace band),
with random directions of the same norm and unrelated word pairs as controls.
Paired counts use exact McNemar tests and trajectory shares Wilcoxon tests, and each claim survives Holm
correction within its own post hoc family, not across claims (Appendix~\ref{app:stats}).

\section{Results}
\label{sec:results}

\subsection{The lens first reads a concept at one attention block}
\label{sec:readable}

\looseness=-1 To find where the lens first reads the instructed hand, we track the mean per-sample cosine of the counterfactual
change $d$ with the side contrast (oriented to the other side). Away from the side word, it stays at most
0.04 through block 14 at the prompt end, in the plan and at the action start (which predicts the first arm token). At attention block 15 the cosine jumps (from 0.02 to 0.17 at the prompt
end), and the largest per-sample step at the prompt end is there in 88\% of the 75. The prompt end precedes every generated token, so this is not a readout of the plan
(Appendix~\ref{app:more}); logit and action lenses agree (Appendix~\ref{app:lens}).
Object nouns step most at block 19 (96\% of 52, Figure~\ref{fig:alignment}).

\subsection{Lens edits change the words and can stall the arm, but do not redirect it}
\label{sec:behaviour}

\paragraph{The J-lens swap redirects no reach.}\looseness=-1 Clamping the side words' lens coordinates in prompt and plan
in generation ($\alpha=1$) flips the plan's side in 88\% of plans that name a side
(ten random draws, 0--6\%), a manipulation check. Yet no reach is redirected (0\%, Figure~\ref{fig:teaser}), while the counterfactual instruction redirects 29\% (Holm
$p=0.037$). The swap can stop reaches (31\% at $\alpha=1$, 51\% at $\alpha=2$), more often than typical
random draws and, at $\alpha=2$, than unrelated word pairs (Appendix~\ref{app:stats}).

\paragraph{The reach comes mainly from the GDN memory.}\looseness=-1 Once the plan's side words and arm tokens are forced
to the other side, the other arm reaches mainly with the counterfactual prompt's GDN memory
(Figure~\ref{fig:teaser}). Forced arm tokens alone leave the other arm idle (17 of 18 well-formed chunks stop).
The other arm reaches in 9\% with forcing alone, in 60\%
with the counterfactual prompt's GDN memory (Holm $p=1.8\times10^{-4}$) and in 17\%
with its attention keys and values (Holm $p=0.0013$ for GDN against attention). A neutral prompt's GDN memory (``free'' or ``bare'' hand) already gives 29\% and 34\%,
though the other side's reproduces more of the motion (Appendix~\ref{app:stats}). Added to the forcing ($\alpha\le4$), the J-lens swap gives 3--14\% reaches with the other arm,
none detectably above forcing alone and all below the GDN memory (Holm $p\le5.8\times10^{-4}$), and its
variants do no better (Appendices~\ref{app:stats},~\ref{app:more}).

\paragraph{The base model follows its plan on a text task.}\looseness=-1 There, Qwen3.5-2B-Base writes a plan line and then a sign or digit codes in \gn{}'s format
(Figure~\ref{fig:contrast}, Appendix~\ref{app:base}). The other side's GDN 10--17 memory flips its plan and sign in
all 400 prompts, yet its output
mostly follows forced plan words or a lens swap. \gn{}'s backbone fails this task, so the comparison crosses tasks and
cannot separate robot training from the robot task.

\subsection{Two routes by block type}
\label{sec:routes}

\looseness=-1 Why do lens edits move the words but not the reach? We patch~\citep{meng2022locating,heimersheim2024how} one block's token-mixing output
(its \emph{mixer}) from the counterfactual into the clean run after the side word,
with the clean plan and arm tokens fixed, and report the share of the effect it reproduces (shares are not
additive; Figure~\ref{fig:patching}).

\looseness=-1 \emph{Attention block 15 carries the words and the lens readout.} It carries 0.44 of the plan-word effect, 0.81 of the lens readout and
0.56 of the arm-token effect, but none of the action-token effect. \emph{The GDN blocks carry the action tokens.} All 18 GDN mixers carry 1.00 of their effect, all 6 attention mixers 0.37 (difference 0.62
[0.50, 0.73]), while the plan word reverses the order (0.84 vs.\ 0.47). On 221 public demonstrations, memory swaps give the same routes, and six random GDN blocks (as many as the attention blocks) carry 0.33 of the action-token effect but 0.09 of the arm-token effect. With flipped instructions alone, plan and arm token disagree in 9\% of 164 runs
(Appendix~\ref{app:demo}).

\looseness=-1 The arm token sits between the routes (Appendix~\ref{app:more}).
The part of $d$ in span($\vv{left},\vv{right}$), which the lens reads, holds 1--5\% of $d$'s energy and at most 0.09 of
the arm-token effect, and steering~\citep{turner2023steering,rimsky2024steering} along it changes the arms in 1 of 36 samples, like a random direction, against 35 along the rest of $d$.

\section{Conclusion}
\label{sec:conclusion}

\looseness=-1 In \gn{}, editing what the J-lens reads drives the words but not the arm, so words and actions share no workspace that the lens can read. Attention
carries the words, and the GDN memory the actions. Steering an embodied model through residual-stream lenses such as the J-lens or logit
lens can therefore miss what it will do. Qwen3.5-2B-Base has this route too, yet follows its plan on a text task.
Whether robot training or the task causes the difference remains open.\label{main-end}%

\section*{Limitations}
We study one model with greedy decoding on 120 observations, with the instructed hand as the main concept. The same
routes appear for object nouns (Appendix~\ref{app:objects}) and, with independent code, on 221 public demonstrations
(Appendix~\ref{app:demo}), but the redirect rates rest on 35 reaching samples, and motion is the decoded action chunk
rather than a trajectory executed on a robot.

On the lens side, the J-lens must be fitted, so what it reads depends on its fitting data. A larger text lens gives the
same directions and results, and the logit lens the same jump and swap counts (Appendix~\ref{app:lens}). Lenses fitted
on action tokens collapse, however, and a state J-lens for the GDN memory fails its positive controls
(Appendix~\ref{app:monitor}), so whether the action can be read from that memory remains open. We also test control only, not whether lens
readouts predict the arm.

As for the interventions, our lens edits are bounded (swaps with $\alpha\le4$ and steering at one block), so larger
edits might still move the arm, although the swap already flips 88\% of plans. The comparison of block types is
confounded with block count, but six random GDN blocks still carry far more of the action-token effect than of the arm-token effect
(Appendix~\ref{app:demo}).

Finally, the base-model control uses one model and one text task, and \gn{}'s backbone cannot do that task, so the
control cannot separate robot training from the robot task. Fine-tuning the base model on an action task in \gn{}'s
format would.

\section*{Ethical Considerations}
Reading or editing an embodied model through what it would say can mislead. In \gn{}, lens edits made the plan and
arm tokens name the other arm without that arm moving, so a monitor of what the model says can misjudge what the robot
will do. Lens edits can also stall a motion, which on hardware could itself be unsafe. We edited a frozen model offline and
executed no motion on a robot. Data and model licenses are listed in Appendix~\ref{app:impl}.

\bibliography{references}

\begin{thebibliography}{42}
\providecommand{\natexlab}[1]{#1}

\bibitem[{Afendulev et~al.(2026)Afendulev, Dontsov, Tutubalina, and
  Korznikov}]{afendulev2026attention}
Kirill Afendulev, Alexey Dontsov, Elena Tutubalina, and Anton Korznikov. 2026.
\newblock \href {https://doi.org/10.18653/2026.findings-emnlp.2061} {What
  attention recalls and recurrence controls in hybrid language models}.
\newblock In \emph{Findings of the Association for Computational Linguistics:
  EMNLP 2026}, pages 43888--43900. Association for Computational Linguistics.

\bibitem[{Baars(1988)}]{baars1988cognitive}
Bernard~J. Baars. 1988.
\newblock \emph{A Cognitive Theory of Consciousness}.
\newblock Cambridge University Press, Cambridge.

\bibitem[{Bandularatne(2026)}]{bandularatne2026jlens}
Nelith Bandularatne. 2026.
\newblock {J-Lens}: A failed replication on {GPT-2}.
\newblock LessWrong, \url{https://www.lesswrong.com/posts/tgn3pD2gLpZvepkWk}.

\bibitem[{Belrose et~al.(2023)Belrose, Ostrovsky, McKinney, Furman, Smith,
  Halawi, Biderman, and Steinhardt}]{belrose2023tuned}
Nora Belrose, Igor Ostrovsky, Lev McKinney, Zach Furman, Logan Smith, Danny
  Halawi, Stella Biderman, and Jacob Steinhardt. 2023.
\newblock \href {https://doi.org/10.48550/arxiv.2303.08112} {Eliciting latent
  predictions from {Transformers} with the tuned lens}.
\newblock \emph{arXiv preprint arXiv:2303.08112}.

\bibitem[{Buurmeijer et~al.(2026)Buurmeijer, Alonso, Swann, and
  Pavone}]{buurmeijer2026observing}
Hugo Buurmeijer, Carmen~Amo Alonso, Aiden Swann, and Marco Pavone. 2026.
\newblock \href {https://doi.org/10.48550/arXiv.2603.05487} {Observing and
  controlling features in {Vision-Language-Action} models}.
\newblock \emph{arXiv preprint arXiv:2603.05487}.

\bibitem[{Chan and Kao(2026)}]{chan2026steering}
Jason Chan and Jonathan~C. Kao. 2026.
\newblock \href {https://doi.org/10.48550/arXiv.2606.15021} {Steering
  autoregressive {Vision-Language-Action} policies via action token
  intervention}.
\newblock \emph{arXiv preprint arXiv:2606.15021}.

\bibitem[{Cheng et~al.(2026)Cheng, Xu, Chen, Chen, Xin, Zhao, Zeng, Liu, Zhang,
  and Jiang}]{cheng2026token}
Ke~Cheng, Xin Xu, Zhichao Chen, Yixiao Chen, Lei Xin, Jianbo Zhao, Fanhu Zeng,
  Yue Liu, Jun Zhang, and Jie Jiang. 2026.
\newblock \href {https://doi.org/10.48550/arXiv.2609.15545} {The token before
  the value is the key}.
\newblock \emph{arXiv preprint arXiv:2609.15545}.

\bibitem[{Dehaene et~al.(1998)Dehaene, Kerszberg, and
  Changeux}]{dehaene1998neuronal}
Stanislas Dehaene, Michel Kerszberg, and Jean-Pierre Changeux. 1998.
\newblock \href {https://doi.org/10.1073/pnas.95.24.14529} {A neuronal model of
  a global workspace in effortful cognitive tasks}.
\newblock \emph{Proceedings of the National Academy of Sciences of the United
  States of America}, 95(24):14529--14534.

\bibitem[{Dettmers et~al.(2022)Dettmers, Lewis, Belkada, and
  Zettlemoyer}]{dettmers2022llmint8}
Tim Dettmers, Mike Lewis, Younes Belkada, and Luke Zettlemoyer. 2022.
\newblock \href {https://doi.org/10.52202/068431-2198} {{LLM.int8()}: 8-bit
  matrix multiplication for transformers at scale}.
\newblock In \emph{Advances in Neural Information Processing Systems 35}, pages
  30318--30332. Neural Information Processing Systems Foundation, Inc.
  (NeurIPS).

\bibitem[{Foutter et~al.(2026)Foutter, Cercola, Wild, Wang, Li, Gammelli, and
  Pavone}]{foutter2026do}
Matthew Foutter, Matteo Cercola, Lena Wild, Yunshan Wang, Michelle Li, Daniele
  Gammelli, and Marco Pavone. 2026.
\newblock \href {https://doi.org/10.48550/arXiv.2607.04681} {Do
  {Vision-Language-Action} models mean what they say? {On} the role of
  faithfulness in embodied reasoning}.
\newblock \emph{arXiv preprint arXiv:2607.04681}.

\bibitem[{Goldowsky-Dill et~al.(2023)Goldowsky-Dill, MacLeod, Sato, and
  Arora}]{goldowskydill2023localizing}
Nicholas Goldowsky-Dill, Chris MacLeod, Lucas Sato, and Aryaman Arora. 2023.
\newblock \href {https://doi.org/10.48550/arXiv.2304.05969} {Localizing model
  behavior with path patching}.
\newblock \emph{arXiv preprint arXiv:2304.05969}.

\bibitem[{Gong et~al.(2026)Gong, Huang, and Wang}]{gong2026first}
Xijie Gong, Zimeng Huang, and Tonghan Wang. 2026.
\newblock \href {https://arxiv.org/abs/2608.31084} {Verbalizing multi-token
  concepts in {LLMs}}.
\newblock \emph{arXiv preprint arXiv:2608.31084}.

\bibitem[{Grant et~al.(2026)Grant, Zhao, and Wang}]{grant2026not}
Bryce Grant, Xijia Zhao, and Peng Wang. 2026.
\newblock \href {https://arxiv.org/abs/2603.19233} {Not all features are
  created equal: A mechanistic study of {Vision-Language-Action} models}.
\newblock In \emph{Advances in Neural Information Processing Systems}.

\bibitem[{Gurnee et~al.(2026)Gurnee, Sofroniew, Pearce, Piotrowski, Kauvar,
  Chen, Soligo, Bogdan, Ong, Wang et~al.}]{gurnee2026workspace}
Wes Gurnee, Nicholas Sofroniew, Adam Pearce, Mateusz Piotrowski, Isaac Kauvar,
  Runjin Chen, Anna Soligo, Paul Bogdan, Euan Ong, Rowan Wang, et~al. 2026.
\newblock \href {https://doi.org/10.48550/arXiv.2607.15495} {Verbalizable
  representations form a global workspace in language models}.
\newblock \emph{arXiv preprint arXiv:2607.15495}.

\bibitem[{H{\"a}on et~al.(2025)H{\"a}on, Stocking, Chuang, and
  Tomlin}]{haon2025mechanistic}
Bear H{\"a}on, Kaylene Stocking, Ian Chuang, and Claire Tomlin. 2025.
\newblock \href {https://arxiv.org/abs/2509.00328} {Mechanistic
  interpretability for steering {Vision-Language-Action} models}.
\newblock In \emph{Proceedings of The 9th Conference on Robot Learning}, volume
  305 of \emph{Proceedings of Machine Learning Research}, pages 2743--2762.
  PMLR.

\bibitem[{He et~al.(2026)He, Shen, Qiu, and Lane}]{he2026sparse}
Matteo He, William~F. Shen, Xinchi Qiu, and Nicholas~D. Lane. 2026.
\newblock \href {https://doi.org/10.48550/arxiv.2609.01936} {Sparse readout
  prism: Explaining {Logit-Lens} scores in features instead of tokens}.
\newblock \emph{arXiv preprint arXiv:2609.01936}.

\bibitem[{Heimersheim and Nanda(2024)}]{heimersheim2024how}
Stefan Heimersheim and Neel Nanda. 2024.
\newblock \href {https://doi.org/10.48550/arXiv.2404.15255} {How to use and
  interpret activation patching}.
\newblock \emph{arXiv preprint arXiv:2404.15255}.

\bibitem[{Jiang et~al.(2026{\natexlab{a}})Jiang, Cao, He, Deng, Wu, Sun, Wang,
  Zhang, Zheng, Song et~al.}]{jiang2026encoded}
Shaohan Jiang, Jiahang Cao, Qiduo He, Fengting Deng, Kun Wu, Jingkai Sun, Jiaxu
  Wang, Qiang Zhang, Qihao Zheng, Chunfeng Song, et~al. 2026{\natexlab{a}}.
\newblock \href {https://arxiv.org/abs/2610.06235} {Encoded but not in control:
  Revealing the grounding gap in {Vision-Language} robot policies}.
\newblock \emph{arXiv preprint arXiv:2610.06235}.

\bibitem[{Jiang et~al.(2026{\natexlab{b}})Jiang, Yuan, Liu, Lu, Cui, Liu,
  Cheng, Gao, Xu, and Zhao}]{galaxea2025g0}
Tao Jiang, Tianyuan Yuan, Yicheng Liu, Chenhao Lu, Jianning Cui, Xiao Liu,
  Shuiqi Cheng, Jiyang Gao, Huazhe Xu, and Hang Zhao. 2026{\natexlab{b}}.
\newblock \href {https://doi.org/10.1109/ICRA57385.2026.11696535} {Galaxea
  open-world dataset and {G0} dual-system {VLA} model}.
\newblock In \emph{2026 IEEE International Conference on Robotics and
  Automation (ICRA)}, pages 20749--20756. IEEE.

\bibitem[{Jin et~al.(2026)Jin, Liang, and Shen}]{jin2026stage}
Baosheng Jin, Yushen Liang, and Hua Shen. 2026.
\newblock \href {https://doi.org/10.18653/2026.findings-emnlp.1148} {{STAGE}:
  Diagnosing semantic transfer at grounded execution in embodied agents}.
\newblock In \emph{Findings of the Association for Computational Linguistics:
  EMNLP 2026}, pages 24309--24331. Association for Computational Linguistics.

\bibitem[{Karnik et~al.(2026)Karnik, Lee, Gupta, and
  Bansal}]{karnik2026reasoning}
Sathwik Karnik, Joseph~JR. Lee, Aryaman Gupta, and Somil Bansal. 2026.
\newblock \href {https://doi.org/10.48550/arxiv.2610.00601} {When reasoning
  helps action: Monitoring and steering chain-of-thought in
  {Vision-Language-Action} policies}.
\newblock \emph{arXiv preprint arXiv:2610.00601}.

\bibitem[{Lanham et~al.(2023)Lanham, Chen, Radhakrishnan, Steiner, Denison,
  Hernandez, Li, Durmus, Hubinger, Kernion et~al.}]{lanham2023measuring}
Tamera Lanham, Anna Chen, Ansh Radhakrishnan, Benoit Steiner, Carson Denison,
  Danny Hernandez, Dustin Li, Esin Durmus, Evan Hubinger, Jackson Kernion,
  et~al. 2023.
\newblock \href {https://doi.org/10.48550/arXiv.2307.13702} {Measuring
  faithfulness in chain-of-thought reasoning}.
\newblock \emph{arXiv preprint arXiv:2307.13702}.

\bibitem[{Liu et~al.(2026)Liu, Dong, Ye, Yuan, Jiang, Yang, Cao, Liu, Sun, Guo
  et~al.}]{galaxea2026g05}
Yicheng Liu, Zibin Dong, Baijun Ye, Tianyuan Yuan, Tao Jiang, Anqi Yang,
  Shicheng Cao, Haonan Liu, Yue Sun, Zihan Guo, et~al. 2026.
\newblock \href {https://arxiv.org/abs/2608.11739} {{G0.5}: One autoregressive
  stream for robot reasoning and action}.
\newblock \emph{arXiv preprint arXiv:2608.11739}.

\bibitem[{Mazaheri(2026)}]{mazaheri2026gathered}
Parsa Mazaheri. 2026.
\newblock \href {https://arxiv.org/abs/2608.15022} {Gathered, not admitted: How
  attention brings a latent variable into verbalizable form}.
\newblock \emph{arXiv preprint arXiv:2608.15022}.

\bibitem[{Meng et~al.(2022)Meng, Bau, Andonian, and
  Belinkov}]{meng2022locating}
Kevin Meng, David Bau, Alex Andonian, and Yonatan Belinkov. 2022.
\newblock \href {https://doi.org/10.52202/068431-1262} {Locating and editing
  factual associations in {GPT}}.
\newblock In \emph{Advances in Neural Information Processing Systems 35}, pages
  17359--17372. Neural Information Processing Systems Foundation, Inc.
  (NeurIPS).

\bibitem[{Merity et~al.(2017)Merity, Xiong, Bradbury, and
  Socher}]{merity2017pointer}
Stephen Merity, Caiming Xiong, James Bradbury, and Richard Socher. 2017.
\newblock \href {https://arxiv.org/abs/1609.07843} {Pointer sentinel mixture
  models}.
\newblock In \emph{International Conference on Learning Representations
  (ICLR)}.

\bibitem[{Nguyen et~al.(2026)Nguyen, Guo, Lampo, Wu, and
  Yaman}]{nguyen2026vladrivebench}
Thach Nguyen, Danhua Guo, Tom Lampo, Fei Wu, and Burhan Yaman. 2026.
\newblock \href {https://doi.org/10.48550/arXiv.2606.12706} {{VLADriveBench}:
  Evaluating {CoT-Action} relationship in {VLA} for autonomous driving}.
\newblock \emph{arXiv preprint arXiv:2606.12706}.

\bibitem[{{nostalgebraist}(2020)}]{nostalgebraist2020logit}
{nostalgebraist}. 2020.
\newblock interpreting {GPT}: the logit lens.
\newblock LessWrong, \url{https://www.lesswrong.com/posts/AcKRB8wDpdaN6v6ru}.

\bibitem[{{Qwen Team}(2026)}]{qwen2026qwen35base}
{Qwen Team}. 2026.
\newblock {Qwen3.5-2B-Base}.
\newblock \url{https://huggingface.co/Qwen/Qwen3.5-2B-Base}.
\newblock Model card; Apache 2.0.

\bibitem[{Rimsky et~al.(2024)Rimsky, Gabrieli, Schulz, Tong, Hubinger, and
  Turner}]{rimsky2024steering}
Nina Rimsky, Nick Gabrieli, Julian Schulz, Meg Tong, Evan Hubinger, and
  Alexander Turner. 2024.
\newblock \href {https://doi.org/10.18653/v1/2024.acl-long.828} {Steering
  {L}lama 2 via contrastive activation addition}.
\newblock In \emph{Proceedings of the 62nd Annual Meeting of the Association
  for Computational Linguistics (Volume 1: Long Papers)}, pages 15504--15522.
  Association for Computational Linguistics.

\bibitem[{Swann et~al.(2026)Swann, McGranahan, Buurmeijer, Kennedy~III, and
  Schwager}]{swann2026sparse}
Aiden Swann, Lachlain McGranahan, Hugo Buurmeijer, Monroe Kennedy~III, and Mac
  Schwager. 2026.
\newblock \href {https://arxiv.org/abs/2603.19183} {Sparse autoencoders reveal
  interpretable and steerable features in {VLA} models}.
\newblock In \emph{Conference on Robot Learning (CoRL)}.

\bibitem[{Timkey and van Schijndel(2021)}]{timkey2021all}
William Timkey and Marten van Schijndel. 2021.
\newblock \href {https://doi.org/10.18653/v1/2021.emnlp-main.372} {All bark and
  no bite: Rogue dimensions in transformer language models obscure
  representational quality}.
\newblock In \emph{Proceedings of the 2021 Conference on Empirical Methods in
  Natural Language Processing}, pages 4527--4546. Association for Computational
  Linguistics.

\bibitem[{Trinh et~al.(2026)Trinh, Azam, Ansari, Ansari, and
  Akhtar}]{trinh2026altered}
Tuan~Duong Trinh, Basim Azam, Mohammed~Ishaq Ansari, Mohammed~Yaqoob Ansari,
  and Naveed Akhtar. 2026.
\newblock \href {https://arxiv.org/abs/2603.12717} {Altered thoughts, altered
  actions: Reasoning chain as control surface for a {Vision-Language-Action}
  policy}.
\newblock \emph{arXiv preprint arXiv:2603.12717}.

\bibitem[{Turner et~al.(2023)Turner, Thiergart, Leech, Udell, Vazquez, Mini,
  and MacDiarmid}]{turner2023steering}
Alexander~Matt Turner, Lisa Thiergart, Gavin Leech, David Udell, Juan~J.
  Vazquez, Ulisse Mini, and Monte MacDiarmid. 2023.
\newblock \href {https://arxiv.org/abs/2308.10248} {Steering language models
  with activation engineering}.
\newblock \emph{arXiv preprint arXiv:2308.10248}.

\bibitem[{Turpin et~al.(2023)Turpin, Michael, Perez, and
  Bowman}]{turpin2023language}
Miles Turpin, Julian Michael, Ethan Perez, and Samuel~R. Bowman. 2023.
\newblock \href {https://doi.org/10.52202/075280-3275} {Language models don't
  always say what they think: Unfaithful explanations in chain-of-thought
  prompting}.
\newblock In \emph{Advances in Neural Information Processing Systems 36}, pages
  74952--74965. Neural Information Processing Systems Foundation, Inc.
  (NeurIPS).

\bibitem[{Wang and Reid(2026{\natexlab{a}})}]{wang2026looped}
Wenlong Wang and Fergal Reid. 2026{\natexlab{a}}.
\newblock \href {https://arxiv.org/abs/2609.01924} {Looped {Transformers} under
  the {Jacobian} lens: Does the global workspace survive recurrence?}
\newblock \emph{arXiv preprint arXiv:2609.01924}.

\bibitem[{Wang and Reid(2026{\natexlab{b}})}]{wang2026residual}
Wenlong Wang and Fergal Reid. 2026{\natexlab{b}}.
\newblock \href {https://arxiv.org/abs/2609.32102} {Residual streams read,
  recurrent states remember: The global workspace in {Mamba} models}.
\newblock \emph{arXiv preprint arXiv:2609.32102}.

\bibitem[{Wu et~al.(2026)Wu, Li, Hermans, Ramos, Bajcsy, and
  Pérez-D'Arpino}]{wu2025do}
Yilin Wu, Anqi Li, Tucker Hermans, Fabio Ramos, Andrea Bajcsy, and Claudia
  Pérez-D'Arpino. 2026.
\newblock \href {https://doi.org/10.1109/ICRA57385.2026.11696392} {Do what you
  say: Steering {Vision-Language-Action} models via runtime reasoning-action
  alignment verification}.
\newblock In \emph{2026 IEEE International Conference on Robotics and
  Automation (ICRA)}, pages 21113--21120. IEEE.

\bibitem[{Yan et~al.(2026)Yan, Ding, Tan, Chen, Wang, Li, and
  Ling}]{yan2026short}
Shi-Qi Yan, Kai-Xuan Ding, Chao-Hong Tan, Qian Chen, Wen Wang, Xiangang Li, and
  Zhen-Hua Ling. 2026.
\newblock \href {https://doi.org/10.48550/arxiv.2608.25347} {Short horizons and
  sparse concepts: A mathematical view of the readout in the {J-lens}}.
\newblock \emph{arXiv preprint arXiv:2608.25347}.

\bibitem[{Yang et~al.(2025)Yang, Kautz, and Hatamizadeh}]{yang2025gated}
Songlin Yang, Jan Kautz, and Ali Hatamizadeh. 2025.
\newblock \href {https://doi.org/10.48550/arXiv.2412.06464} {{Gated Delta
  Networks}: Improving {Mamba2} with delta rule}.
\newblock In \emph{International Conference on Learning Representations}.

\bibitem[{Yin et~al.(2026)Yin, Lin, Xu, Tam, Zeng, Liu, and
  Yin}]{yin2025deepthinkvla}
Cheng Yin, Yankai Lin, Wang Xu, Sikyuen Tam, Xiangrui Zeng, Zhiyuan Liu, and
  Zhouping Yin. 2026.
\newblock \href {https://arxiv.org/abs/2511.15669} {{DeepThinkVLA}: Enhancing
  reasoning capability of {Vision-Language-Action} models}.
\newblock In \emph{Third Conference on Language Modeling (COLM)}.

\bibitem[{Zawalski et~al.(2025)Zawalski, Chen, Pertsch, Mees, Finn, and
  Levine}]{zawalski2024robotic}
Micha{\l} Zawalski, William Chen, Karl Pertsch, Oier Mees, Chelsea Finn, and
  Sergey Levine. 2025.
\newblock \href {https://arxiv.org/abs/2407.08693} {Robotic control via
  embodied chain-of-thought reasoning}.
\newblock In \emph{Proceedings of The 8th Conference on Robot Learning}, volume
  270 of \emph{Proceedings of Machine Learning Research}, pages 3157--3181.
  PMLR.

\end{thebibliography}

\appendix
\section{Statistics}
\label{app:stats}

\paragraph{Rate definitions.} An arm's largest joint range is the maximum over its joints of max minus min over the decoded chunk. Redirect
and stop rates are over the 35 reaching samples, and rates of flipped plans over the
51 plans that name a side (\S\ref{sec:setup}). A plan is \emph{flipped} when its first side word names the other side,
and the 44 plans that name a hand give the rates for the first arm token. A chunk is \emph{well formed} when it has
exactly one arm token per level per arm. Steering is scored by whether the set of moving arms changes, over the 36 samples
whose clean run moves an arm (the 35 and one sample that moves 0.26\,rad).

\paragraph{Statistical tests.} All conditions run on the same samples, so we compare paired counts with exact McNemar tests, give
differences of rates Newcombe's hybrid score 95\% intervals, and Holm-adjust the $p$-values within one family per claim.
The families are the 24 key contrasts, the steering grid (30 tests) and one family for each follow-up claim. We defined
them after exploratory analyses of the same 75 samples, without held-out confirmation or correction across families, so
all $p$-values are post hoc.

\paragraph{Random controls and effect shares.} Random controls are single draws (seed 0 for the swap, one seed per sample for
steering), except that the unforced swap of the side words is compared with ten draws. The trajectory share
$r = \langle D, D_{\text{ref}}\rangle / \|D_{\text{ref}}\|^2$ measures how much of the counterfactual motion an
intervention reproduces, where $D$ is a run's joint trajectory minus the clean one and $D_{\text{ref}}$ the same for the
counterfactual prompt. We compare shares with Wilcoxon signed-rank tests over the samples whose reference change exceeds
0.3\,rad. Patching shares are ratios of sums over samples, with 1000-resample bootstrap intervals.

\begin{table}[ht]
  \centering\small
  \setlength{\tabcolsep}{3pt}
  \begin{tabular}{@{}lr@{}}
    \toprule
    condition (outcome) & $k/n$ [Wilson 95\%] \\
    \midrule
    clean: reaching arm is instructed & 35/35 [0.90, 1.00] \\
    cf.\ instruction (plan flipped) & 30/51 [0.45, 0.71] \\
    cf.\ instruction (redirected) & 10/35 [0.16, 0.45] \\
    J-lens swap, $\alpha{=}1$ (plan flipped) & 45/51 [0.77, 0.94] \\
    \quad ten random draws & 0--3 (median 1) \\
    J-lens swap, $\alpha{=}1$ (redirected) & 0/35 [0.00, 0.10] \\
    J-lens swap, $\alpha{=}1$ (stopped) & 11/35 [0.19, 0.48] \\
    \quad ten random draws & 0--5 (median 4) \\
    J-lens swap, $\alpha{=}2$ (stopped) & 18/35 [0.36, 0.67] \\
    \quad ten random draws & 4--11 (median 7) \\
    \quad ``up''/``down'', ``day''/``night'' & 1/35, 3/35 \\
    forced plan words (redirected) & 2/35 [0.02, 0.19] \\
    forced words + arm tokens (redirected) & 3/35 [0.03, 0.22] \\
    \quad + J-lens swap, $\alpha{=}1$ & 3/35 [0.03, 0.22] \\
    \quad + J-lens swap, $\alpha{=}2$ & 1/35 [0.01, 0.15] \\
    \quad + J-lens swap, $\alpha{=}4$ & 3/35 [0.03, 0.22] \\
    \quad\quad prompt only & 5/35 [0.06, 0.29] \\
    \quad + GDN memory (18 blocks) & 21/35 [0.44, 0.74] \\
    \quad + ``free'' prompt's GDN memory & 10/35 [0.16, 0.45] \\
    \quad + ``bare'' prompt's GDN memory & 12/35 [0.21, 0.51] \\
    \quad + 6 attention blocks' K/V & 6/35 [0.08, 0.33] \\
    \bottomrule
  \end{tabular}
  \caption{Key counts with Wilson 95\% intervals. ``cf.'' is the counterfactual instruction in free generation, and
  the J-lens swap edits the prompt and plan. The main text and the paragraphs below give the paired tests.}
  \label{tab:counts}
\end{table}

\paragraph{Contrasts that hold after Holm correction.} The counterfactual instruction redirects more reaches than the
$\alpha=1$ J-lens swap (all 10 discordant pairs, $p=0.002$) and flips fewer plans (3 vs.\ 18, $p=0.0015$), which gives Holm $p=0.037$ and 0.030 with the 24 key contrasts. With forced words and arm tokens, the directed clamp redirects
fewer reaches than the GDN memory (4--6 vs.\ 21 of 35, Holm $p\le0.0041$), and the other side's GDN memory moves the
other arm further than a neutral prompt's ($r$ larger in 21 of 27 samples, Holm $p=0.005$). Without forcing, the swap
stops more reaches than random edits pooled over ten draws (Holm $p\le1.8\times10^{-5}$) and, at $\alpha=2$, than swaps
of unrelated word pairs (Holm $p\le0.001$), although their edits are 12--28\% smaller. Steering along the part of the
counterfactual change that the lens does not read moves the arms far more often than along the part that it reads
(Holm $p=3.4\times10^{-9}$), which does no better than a random direction.

\paragraph{Joint outcomes and reach threshold (post hoc).} Reclassifying the saved runs, which reproduces all 19
counts of Table~\ref{tab:counts}: of the 21 reaching samples whose clean plan names a side, the $\alpha=1$ swap flips the
plan in 17, of which 8 still reach with the instructed arm, 9 stop and none reach with the other arm ($\alpha=2$: 15
flipped, 5 reach, 9 stop, 1 moves both arms).
In these 17 flipped plans, the first arm token flips in 4, all of which stop, and stays on the instructed side in the other 13
(8 reach with the instructed arm, 5 stop).
For reach thresholds from 0.1 to 0.5\,rad, the swap redirects 0 of 35, and the counterfactual instruction vs.\ the swap,
forcing with vs.\ without the GDN memory, and attention keys and values vs.\ the GDN memory keep all discordant pairs in
one direction (Holm $p\le0.002$). The other arm's joint range minus the instructed arm's orders them the same way
(Wilcoxon, Holm $p\le7.1\times10^{-5}$). Counting malformed chunks as non-reaches turns 21 into 20 and 3 into 2 of 35.

\paragraph{Contrasts that do not hold.} Against single random draws, the stalls are less clear. At $\alpha=2$ the
swap stops 18 reaches against 11 for the least favorable draw ($p=0.092$). Lens edits added to the forcing change no
redirects detectably. No J-lens swap or directed condition at $\alpha\le4$ differs from forcing alone ($p\ge0.25$, with
a 95\% upper bound of about 7 of 35 reaches), and none differs from the neutral prompts' memory alone ($p\ge0.18$).

\section{Additional results for the hand}
\label{app:more}

\begin{figure}[t]
  \centering
  \includegraphics{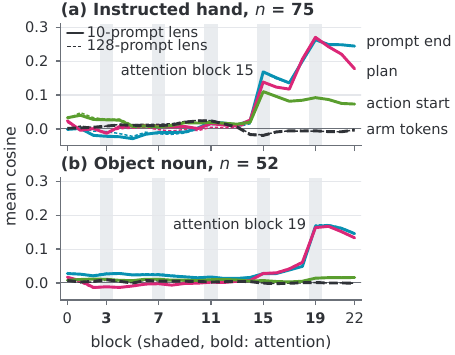}
  \caption{\textbf{The lens first reads a concept at one attention block.} Mean per-sample cosine of the
  counterfactual change $d$ with $v_{\text{target}} - v_{\text{source}}$, by block and position: (a)~hand;
  (b)~object nouns. Line colors mark positions only, not the arm sides or routes of Figures~\ref{fig:teaser}
  and~\ref{fig:pair}.}
  \label{fig:alignment}
\end{figure}

\paragraph{First lens readout of the hand.} At the prompt end, the mean cosine of $d$ with the side contrast
jumps from 0.024 to 0.17 at block 15 with either lens (Figure~\ref{fig:alignment}a), and the largest per-sample step falls
there in 66 of 75 samples. The tokens right after the side word align from block 4 already, so block 15 is where the hand
first reaches distant positions. In the plan, the step is less concentrated (block 15 in 42 of 75 samples), and
alignment rises again at blocks 18--19.

\paragraph{Patching shares with intervals.} The counterfactual raises the other arm's token by 1.56 logits and the
other side's plan word by 10.3, and it lowers the log-probability of the clean action tokens by 0.065 nats per token.
Table~\ref{tab:patching} gives the shares of these effects that patching reproduces. With block 15's output held clean by path
patching~\citep{goldowskydill2023localizing}, all GDN mixers carry 0.42 instead of 0.90 of the arm-token effect, and the
part of $d$ that the lens reads carries 0.13 of it with the logit lens. The GDN mixers keep 0.84 [0.76, 0.93] of their action-token share but only 0.47 [0.35, 0.57] of their arm-token share, so part of
their arm-token effect passes through block 15. No block before block 10 carries more than 0.035 of any effect.

\begin{table}[t]
  \centering\small
  \begin{tabular}{@{}llr@{}}
    \toprule
    patched mixers & output & share [95\% interval] \\
    \midrule
    block 15 & plan word & 0.44 [0.40, 0.48] \\
     & lens readout & 0.81 [0.73, 0.90] \\
     & arm token & 0.56 [0.47, 0.65] \\
     & action tokens & $-0.01$ [$-0.05$, 0.02] \\
    all 18 GDN & plan word & 0.47 [0.41, 0.53] \\
     & action tokens & 1.00 [0.99, 1.00] \\
    all 6 attention & plan word & 0.84 [0.81, 0.87] \\
     & action tokens & 0.37 [0.27, 0.49] \\
    \bottomrule
  \end{tabular}
  \caption{Share of the counterfactual effect that patching reproduces, with bootstrap 95\% intervals over 75 samples.}
  \label{tab:patching}
\end{table}

\paragraph{Forced arm tokens.} Forced arm tokens alone leave the other arm idle (17 of 18 well-formed chunks stop), so
with the clean memory an arm token selects which arm's action tokens are emitted, not their content.

\paragraph{Lens swap variants.} No variant of the lens edit redirects detectably more reaches. A swap along logit-lens
directions matches the J-lens swap on every outcome (paired $p\ge0.38$). A swap limited to blocks 15--21 edits the plan
more but redirects no reach without forcing, and with forced words and arm tokens it redirects about as many as forcing
alone (2--4 vs.\ 3 of 35). The \emph{directed clamp}, which edits only the rows in which the unedited run reads the
instructed side above the other, redirects far fewer than the GDN memory, and adding either edit to a neutral prompt's
GDN memory changes no redirects detectably (Appendix~\ref{app:stats}).

\paragraph{Part of the change that the lens reads.} If the lens read the signal that selects the arm, the part of $d$ in
span($\vv{left},\vv{right}$) would carry much of the arm-token effect, but it carries little (\S\ref{sec:routes}). Steering
shows the same split with a larger edit. We add 16 times a direction to block 18's output at the action start. Along the
mean counterfactual change and along its part orthogonal to the lens span, the moving arms change in 35 of 36 samples.
Along the part that the lens reads, they change in only 1, as along a random direction of the same norm.

\section{Object nouns}
\label{app:objects}

To check that the routes are not specific to the hand, we repeat the analysis on the 52 samples whose instruction
and clean plan share a single-token object noun. The noun is replaced by a partner from the same category under three
rules (the \emph{next} noun, the next \emph{distinct} one, or the \emph{second} after it). Under every rule, the lens first
reads the object at attention block 19, where the largest cosine step falls in 45--51 of 52 samples
(Figure~\ref{fig:alignment}b), as it reads the hand at block 15. As for the hand, all GDN mixers carry all of the action-token
effect (1.00, against 0.17--0.25 for all attention mixers), while all attention mixers carry 0.81--0.82 of the effect on
the plan noun, most of it at block 19 (0.36--0.43). The action-token effect is small (median magnitude at most 0.016 nats per
token), but wherever it reaches 0.02 nats per token, the GDN mixers carry at least 0.8 of it.

\section{Choice of lens}
\label{app:lens}

\paragraph{Lens fitted on 128 prompts.} The main lens is fitted on 10 prompts, which suffice in
\citet[App.~A.7]{gurnee2026workspace}, so we also fit a lens on 128 WikiText-103 records that include them. It reads no
better (top-1 agreement with the model's final token at block 18 of 0.144, against 0.159 for the main lens) and gives
the same directions. In blocks 9--21, the side contrast has a cosine of 0.96--1.00 between the two lenses and of
0.93--1.00 between the main lens and lenses fitted on three disjoint prompt sets. With the larger lens, the hand's
block-15 step and the J-lens swap change by at most 3 samples, although single top-1 readouts at \gn{}'s positions agree
on only 37--70\%. We did not repeat the patching and steering with it.

\paragraph{Failure of action lenses.} Lenses fitted on 10 or 100 of the policy's own generations read \texttt{"} at
67--100\% of positions, because residual dimension 1201 is large and negative at every position. Such outlier
dimensions~\citep{timkey2021all,dettmers2022llmint8} also break a sparsified GPT-2 J-lens~\citep{bandularatne2026jlens}.
With dimension 1201 zeroed, the 100-sample action lens shows the block-15 step of the text and logit lenses (a per-sample
jump of 0.146 at the prompt end, against 0.144 for both), so the step is a property of the residual stream, not of the
J-lens alone. We did not run the swap with this lens.

\section{The base model on a text task}
\label{app:base}

\paragraph{Model choice.} To ask whether the split exists before robot training, we use
Qwen3.5-2B-Base~\citep{qwen2026qwen35base}, which has \gn{}'s backbone architecture (the same 24 blocks, attention
blocks and widths) but no robot training. \gn{}'s report says that it was initialized from Qwen3.5-2B (\S\ref{sec:setup}),
probably from this base release, since its processor path names a base processor, but we did not compare weights.

\paragraph{Text task design.} Without action tokens, the base model gets a text task with the structure of \gn{}'s output, a
written plan followed by the action. Each prompt has four demonstrations and then a query of the form
\begin{quote}\small\ttfamily\raggedright\frenchspacing
Task: Move the marker 2 squares to the left. State: the marker is on square 7; the squares are numbered 0 to 9; the
board is green.\\
Plan: move the marker left.\\
Move: -2\\
Path: 6,5
\end{quote}
The model writes the plan line, then the sign (`` -'' or `` +''), which we score as the output that follows the plan,
then the magnitude and the path. A \emph{pair} is two prompts that differ only in the query's side word. In the
\emph{mid} set, 27 tokens follow the side word, within \gn{}'s range of 19--37, and the model gets the plan, the sign and
the magnitude right in both prompts of all 400 mid pairs. An undefined outcome counts as not flipped.

\paragraph{Forced plan line.} With the plan line forced to the side that the prompt does not name, in both directions
of every pair, the sign follows the plan in 716 of 800 trials, so the base model acts on its written plan.

\paragraph{Pilot J-lens swap.} With a J-lens fitted by \gn{}'s recipe on 32 WikiText records, the swap of
\S\ref{sec:setup} at blocks 9--21 flips the plan word in all 100 prompts of 50 mid pairs and the sign in 95 (90 with the
logit lens), while none of 24 random controls with larger edits flips either. Unlike in \gn{}, editing what the lens reads
thus moves the output. As a 50-pair pilot, this result is exploratory.

\paragraph{Memory swaps.} As in \gn{} (\S\ref{sec:behaviour}), each prompt is decoded with part of its cache, the GDN
states or the attention keys and values, taken from the other prompt of its pair. The memory of GDN blocks 10, 12--14,
16 and 17 flips both the plan and the sign in all 400 mid prompts, while that of the 6 attention blocks flips none, so
the recurrent route exists before robot training.

\paragraph{Output in \gn{}'s format.} \gn{}'s output form, an arm token followed by a chunk of action tokens, might cause the
dissociation by itself. We therefore give the first 100 mid pairs an output of that form, the plan line, then ``Move:''
with a sign (`` -'' or `` +'') in place of the arm token and eight digit codes that give the marker's square at eight ticks (``Codes: 7,6,5,5,5,5,5,5''
for the prompt above). With the plan forced to the other side, the digit codes follow it in 192 of 200 prompts, above the
pre-declared bar of 0.80 for no dissociation.
 With plan and sign both forced, they follow in 199 of 200, and in all
200 with the other prompt's GDN memory as well (Figure~\ref{fig:contrast}), while \gn{}'s other arm reaches in only 3 of
35 samples. The output form alone thus does not produce the dissociation, although with the side carried by one code,
this test rules out only this simple form.

\paragraph{\gn{}'s backbone on the text task.} The cleanest test would run \gn{} on this task, but its backbone,
converted bit for bit to the base model's format, gets no pair right among the first 50 of any set, as it lacks the base
model's few-shot behavior on text. The comparison in \S\ref{sec:behaviour} therefore crosses tasks.

\section{A state J-lens for the GDN memory}
\label{app:monitor}

Because the residual-stream J-lens cannot read the GDN memory, we fit the full-path and readout-only state J-lenses of
\citet{wang2026residual} to the recurrent states of GDN blocks 10, 12--14, 16 and 17. Both pass their consistency checks
but fail both pre-declared positive controls. They read the current token at a top-10 rate of at most 0.091, below a
constant mean-logit readout in 5 of 6 blocks, while the residual J-lens reaches 0.27--0.66. They also rank the side word
present in the prompt above the absent one in only 60--85 of 150 cases, near the 75 of a lens that ignores the prompt
and far below the pre-declared bar of 100. This lens thus cannot tell what the memory holds.

\section{Results on public demonstrations}
\label{app:demo}

\paragraph{Episodes and outcome measures.} To test whether the routes depend on our observations and our code, an independent analysis
uses 221 public demonstration episodes that Galaxea released with its G0 models (two pick-and-place tasks, CC BY-NC-SA
4.0). It decodes the first frame of each episode greedily, once with its instruction and once with the side word flipped
(442 runs). It scores the first arm token by $A$, the logit of \tok{<left_control_0>} minus that of \tok{<right_control_0>},
and the action tokens by their log-probability under the clean plan. Where the plan names a side (164 runs), the arm token
disagrees with it in 15, all with a flipped instruction and none with the original one, so plan and first arm token can
diverge without any internal edit.

\paragraph{Memory swap results.} With the clean plan forced, we copy parts of the memory at the prompt end from the flipped prefill and
take the median share of the change. The arm token draws mainly on attention, whose keys and values carry 0.52 of the change
in $A$ (block 15 alone 0.47) against 0.35 for the GDN recurrent states. The action tokens draw on the GDN memory, which carries
0.97 of their change against 0.10 for the attention cache. Six random GDN blocks, as many as there are attention blocks,
carry 0.09 of $A$ but 0.33 of the action tokens' change, so the split does not follow block count. Unlike in \S\ref{sec:routes}, the
memory of GDN block 1 alone carries 0.22 of the action tokens' change, which our patching of mixer outputs after the side word does not
show, as the two interventions edit different things.

\paragraph{Limits of this sample.} This sample measures the first arm token, not motion. It runs in BF16 without Holm correction, and its
episodes may overlap the data on which Galaxea trained its models.

\section{Licenses and compute}
\label{app:impl}

We do not redistribute the model weights or the robot recordings, which are third-party assets. The observations
come from the Galaxea Open-World Dataset and, in Appendix~\ref{app:demo}, Galaxea's public G0 demonstrations (both CC BY-NC-SA 4.0). We use \gn{} under the G0.5 Community License (non-commercial), and the base-model control
(Appendix~\ref{app:base}) uses Qwen3.5-2B-Base~\citep{qwen2026qwen35base} under the Apache 2.0 license. Text lenses are fitted on
WikiText-103~\citep{merity2017pointer} (CC BY-SA). We used all of them for non-commercial research only. We will release the code, the sample manifest, the model,
codec and tokenizer versions, the prompts, the forcing rules, per-condition outputs and the scripts for every table and
figure.

Each run used one 48\,GB NVIDIA RTX A6000 GPU, except those of Appendices~\ref{app:base} and~\ref{app:monitor} (one NVIDIA L40S, and one NVIDIA H100 for the state J-lens) and of Appendix~\ref{app:demo} (NVIDIA L4 GPUs and one RTX 5070 Ti). The
128-prompt text lens took 4.2\,h on a shared GPU and the 10- and 100-sample action lenses 0.6\,h and 6.1\,h. Statistics and figures run on a CPU from the saved outputs.
We used AI assistants for coding and for editing the text. The authors conceived and designed the study, made all
analysis and framing decisions, and checked every reported number against the saved outputs.

\end{document}